\documentclass[conference]{IEEEtran}
\IEEEoverridecommandlockouts

\usepackage{multirow}
\usepackage{hhline}
\usepackage{verbatim}
\usepackage{cite}
\usepackage{enumitem}
\usepackage{algorithm}
\usepackage{algpseudocode}
\usepackage{graphics} 
\usepackage{graphicx}
\usepackage{subfig}
\usepackage{url}
\usepackage{balance}
\usepackage{siunitx}
\usepackage{booktabs}
\usepackage[font={small}]{caption}
\usepackage{amsmath,amssymb,amsfonts}

\usepackage{hyperref} 
\begin{document}
\vspace{-3mm}
\title{Global Cross-Time Attention Fusion for Enhanced Solar Flare Prediction from Multivariate Time Series}

\author{\IEEEauthorblockN{Anonymous}}

\author{
    \IEEEauthorblockN{Onur Vural, Shah Muhammad Hamdi, Soukaina Filali Boubrahimi}
    \IEEEauthorblockA{Department of Computer Science, Utah State University, Logan, UT, USA\\
    Email: a02429549@aggies.usu.edu, hamdi@usu.edu, boubrahimi@usu.edu}
}

\author{
    \IEEEauthorblockN{Onur Vural, Shah Muhammad Hamdi, Soukaina Filali Boubrahimi}
    \IEEEauthorblockA{\textit{Department of Computer Science}, \textit{Utah State University}, Logan, UT 84322, USA\\
    Emails: \{onur.vural, s.hamdi, soukaina.boubrahimi\}@usu.edu}
}

\maketitle

\begin{abstract}
Multivariate time series classification is increasingly investigated in space weather research as a means to predict intense solar flare events, which can cause widespread disruptions across modern technological systems. Magnetic field measurements of solar active regions are converted into structured multivariate time series, enabling predictive modeling across segmented observation windows. However, the inherently imbalanced nature of solar flare occurrences, where intense flares are rare compared to minor flare events, presents a significant barrier to effective learning. To address this challenge, we propose a novel Global Cross-Time Attention Fusion (GCTAF) architecture, a transformer-based model to enhance long-range temporal modeling. Unlike traditional self-attention mechanisms that rely solely on local interactions within time series, GCTAF injects a set of learnable cross-attentive global tokens that summarize salient temporal patterns across the entire sequence. These tokens are refined through cross-attention with the input sequence and fused back into the temporal representation, enabling the model to identify globally significant, non-contiguous time points that are critical for flare prediction. This mechanism functions as a dynamic, attention-driven temporal summarizer that augments the model’s capacity to capture discriminative flare-related dynamics. We evaluate our approach on the benchmark solar flare dataset and show that GCTAF effectively detects intense flares and improves predictive performance, demonstrating that refining transformer-based architectures presents a high-potential alternative for performing solar flare prediction tasks.
\end{abstract}


\section{Introduction} \label{sec:intro}

An intense solar flare erupted from the Sun's surface in 1859, unleashing a powerful geomagnetic storm that disrupted telegraph systems across Europe and North America, sparked widespread auroral displays visible near the equator, and induced electrical currents strong enough to ignite fires in communication lines \cite{giegengack2015carrington}. This occurrence, known as the Carrington Event, is a striking and extreme historical example of the effects of an intense solar flare~\cite{angryk2020multivariate}. Solar flares are explosive manifestations of solar magnetic energy, erupting from active regions on the Sun’s surface and extending into the corona and interplanetary space~\cite{mcintosh1990classification}. These bursts are classified on a logarithmic scale as A, B, C, M, and X, based on peak soft X-ray intensity in the 1 to 8 Ångström band, with M and X flares indicating intense flare events~\cite{ahmadzadeh2021train}. Beyond their astrophysical intrigue, such eruptions bear serious consequences: they unleash radiation across the electromagnetic spectrum, including gamma rays and extreme ultraviolet, that can threaten astronaut safety, compromise satellites and power grids, and disrupt navigation and communications infrastructures. Furthermore, an occurrence of similar scale to the Carrington Event could lead to devastating societal and economic consequences in our technology-driven world~\cite{eastwood2017economic}. In light of these risks, the scientific community continues to advance solar flare prediction by analyzing solar magnetic field data and developing predictive models.

Magnetic field measurements from solar active region vector magnetogram data have been increasingly organized into structured multivariate time series (MVTS) over segmented time windows, enabling MVTS-driven predictive modeling that outperforms single-timestamp approaches~\cite{angryk2020multivariate}. However, severe class imbalance due to rarity of intense flares is a central challenge, biasing models to favor the majority class~\cite{hamdi2017time}. Traditional classifiers and sequence models typically exploit only temporal dependencies in MVTS data, limiting capture of complex inter-variable patterns and resulting in suboptimal performance in identifying intense flaring events. To address these limitations, transformer-based architectures emerge as promising candidates, utilizing self-attention mechanisms to dynamically weigh the relevance of each time step and feature in a sequence, capturing both local and global dependencies without the sequential limitations of traditional models. In solar flare prediction, where critical magnetic patterns may be subtle and dispersed across time, such expressiveness is particularly advantageous ~\cite{alshammari2024transformer}. Additionally, cross-attention mechanisms, which relate different modalities or segments of data, can further help to identify rare but significant flares.

In this work, we introduce a transformer-based classification framework, Global Cross-Time Attention Fusion (GCTAF), tailored to improve the prediction of solar flare events from MVTS data. Unlike generic transformer models that rely solely on self-attention over local temporal interactions, GCTAF introduces learnable global tokens that attend to the entire input sequence via a cross-attention mechanism. This enables the model to capture globally salient temporal dependencies that conventional self-attention may overlook. These global representations are fused with local temporal features extracted through stacked transformer encoder modules, allowing the model to integrate both fine-grained and holistic information. GCTAF is assessed using the MVTS solar flare benchmark dataset, where enhanced prediction performance indicates higher effectiveness in identifying intense flares. These findings underscore the efficacy of refined transformer models in addressing the inherent challenges of solar flare prediction.

\section{Related Work} \label{sec:Related Work}

\subsection{A Brief History of Solar Flare Prediction}
Early solar flare prediction relied on expert systems grounded in sunspot classification \cite{mcintosh1990classification}. With advances in observational technology, magnetic field data became increasingly accessible, shifting prediction toward data-driven approaches \cite{vural2024contrastive}. NASA’s Solar Dynamics Observatory (SDO) mission, particularly its Helioseismic and Magnetic Imager (HMI) instrument, enabled large-scale, high-cadence magnetogram data analysis \cite{ahmadzadeh2021train}. Consequently, flare prediction evolved into a machine learning task employing traditional classifiers such as SVMs \cite{bobra2015solar}, and logistic regression \cite{song2009statistical}. Building on these efforts, the Space Weather Analytics for Solar Flares (SWAN-SF) dataset was introduced \cite{angryk2020multivariate}, capturing magnetic field evolution as MVTS data. Since then, MVTS-based approaches have become central to advancing solar flare prediction. These include preprocessing-enhanced decision trees \cite{ma2017solar}, contrastive representation learning to emphasize class distinctions \cite{vural2024excon, vural2024contrastive, vural2025contrastive}, deep LSTM architectures for temporal sequence modeling \cite{muzaheed2021sequence}, and graph-based methods \cite{vural2025solar}.

\subsection{Transformer-Based Time Series Classification}

The emergence of transformers represents a breakthrough in deep learning approaches by employing self-attention mechanisms to capture long-range dependencies in sequential data without relying on recurrent or convolutional operations \cite{vaswani2017attention}. This innovation enabled parallel processing and greater flexibility in modeling sequence data, which has since propelled transformers beyond natural language processing into diverse domains, including time series classification. In time series tasks, transformers have shown strong potential by learning contextual temporal features across multiple channels while accommodating variable-length sequences \cite{wen2022transformers}. Selected examples of recent advancements include transformer-based unsupervised MVTS representation learning for improved downstream classification \cite{zerveas2021transformer}, gated transformer architectures for enhanced feature extraction \cite{liu2021gated}, shapelet-based modules to capture both class-specific and general patterns \cite{le2024shapeformer}, and the use of hierarchical pooling, adaptive frequency learning strategies, and feature-map-wise attention mechanisms to model multi-scale temporal dependencies \cite{yang2024dyformer}. Collectively, these developments emphasize the transformer’s versatility and growing impact in time series classification research.

\section{Dataset} \label{sec:Dataset}

Although machine learning has been applied to solar flare prediction, earlier efforts primarily relied on static, point-in-time observations \cite{bobra2015solar}. To address this limitation, the SWAN-SF dataset was introduced \cite{angryk2020multivariate}, providing a temporally resolved framework that captures the dynamic evolution of solar magnetic fields through MVTS data. Each SWAN-SF instance is an MVTS segment $mvts^{(k)} \in \mathbb{R}^{\tau \times N}$, where $\tau$ is the sliding step and $N$ represents 24 magnetic field parameters \cite{bobra2015solar}. Segments are extracted with a sliding window of $T_{\text{obs}} = 12$ hours, prediction window $T_{\text{pred}} = 24$ hours, and step size $\tau = 1$ hour \cite{ahmadzadeh2021train, angryk2020multivariate}. Each is labeled according to the most intense flare in the subsequent prediction window. SWAN-SF includes five flare categories—FQ, B, C, M, and X—consolidated into non-flare (NF = FQ, B, C) and flare (F = M, X) classes for binary flare prediction tasks. SWAN-SF has a chronic, severe imbalance between F and NF classes due to the rarity of M- and X-category events. This imbalance poses a difficulty in building robust flare classification systems, leading models to favor the extremely prevalent NF class \cite{hamdi2017time, bobra2015solar, vural2025solar}. Table~\ref{tab:flare_classes} provides a summary of flare categories, their respective peak flux thresholds, and their proportion within SWAN-SF.

\begin{table}[h]
\footnotesize
\centering
\caption{Flare categories, peak flux ranges, and dataset proportions.}
\setlength{\tabcolsep}{3pt} 
\begin{tabular}{llcc}
\toprule
\textbf{Class Label} & \textbf{Flare Category} & \textbf{Peak Flux (W/m\textsuperscript{2})} & \textbf{Percentage (\%)} \\
\midrule
\multirow{4}{*}{Non-Flare (NF)} 
& Flare-Q \& A (FQ) & $<$ $10^{-7}$ & 82.8\% \\
& B & $10^{-7}$ to $10^{-6}$ & 5.5\% \\
& C & $10^{-6}$ to $10^{-5}$ & 9.8\% \\
\midrule
\multirow{2}{*}{Flare (F)} 
& M & $10^{-5}$ to $10^{-4}$ & 1.7\% \\
& X & $>$ $10^{-4}$ & 0.16\% \\
\bottomrule
\end{tabular}
\label{tab:flare_classes}
\end{table}

\section{Methodology} \label{sec:Methodology}

\begin{figure*}[ht]
\centering
\includegraphics[width = 0.90\linewidth]{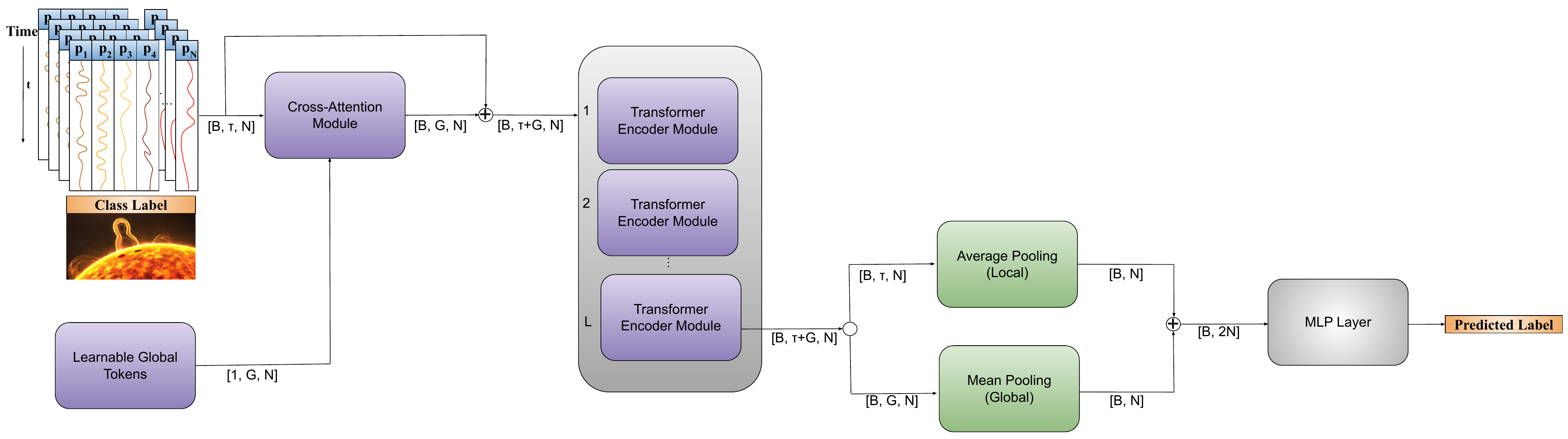}
\caption{The GCTAF model for solar flare prediction takes input of shape \([B, \tau, N]\) and learnable global tokens \([1, G, N]\) shared across batches. The global tokens attend to the input via cross-attention, producing \([B, G, N]\). The input and global tokens are concatenated to \([B, \tau+G, N]\) and processed by transformer encoder blocks. The output is split into local \([B, \tau, N]\) and global \([B, G, N]\) tokens. Local tokens are pooled to \([B, N]\), global tokens averaged to \([B, N]\), then concatenated into \([B, 2N]\) and passed through an MLP for final logits.}
\label{fig:GCTAF_arch}
\end{figure*}

GCTAF framework comprises multiple components that combine global and local temporal information from MVTS data by leveraging transformer-based strategies. Fig.~\ref{fig:GCTAF_arch} provides an overview of the proposed framework architecture. GCTAF attends learnable global tokens to input sequences via cross-attention. The fused representation is refined through transformer-based modules, pooled, and classified using an MLP head. The details of the transformer-based module as the encoder, and overall framework architecture are discussed in following sections respectively.

\subsection{Transformer Encoder Module}

The transformer-based encoder module serves as the core temporal modeling unit, designed to capture complex temporal dependencies and interactions among solar magnetic field parameters \cite{vaswani2017attention, alshammari2024transformer}. By employing self-attention and non-linear transformations within a residual framework, it seeks to obtain representations emphasizing salient temporal regions while suppressing irrelevant noise, enabling effective flare prediction. The module consists of the following components:

\begin{itemize}[leftmargin=*]
    \item \textbf{Layer normalization for attention:} Applies normalization over the embedding dimension to stabilize training and decrease internal covariate shift.

    \item \textbf{Multi-head self-attention (MHSA):} Computes attention weights by projecting the normalized input into queries, keys, and values through learned linear transformations. Multiple attention heads ($h$) allow the model to jointly attend to information from various temporal subspaces \cite{alshammari2024transformer}.

    \item \textbf{Dropout after attention:} Dropout is applied to the attention output with a fixed probability to prevent co-adaptation of attention heads and mitigate overfitting, particularly important due to the limited F-class samples compared to NF-class.

    \item \textbf{Layer normalization for feedforward network:} Another normalization layer is applied before the feedforward network, ensuring a stable input distribution across layers.

    \item \textbf{Feedforward network (FFN):} A two-layer position-wise feedforward subnetwork applies the same transformation to each time step independently. It consists of a linear projection from embedding dimension $N$ to a higher-dimensional space, followed by a ReLU activation, a dropout layer for regularization, and another linear layer projecting back to $N$ \cite{vaswani2017attention}. This enhances representational capacity to model complex, non-linear relationships among solar indicators.

    \item \textbf{Residual Connections:} Both attention and FFN sublayers are equipped with skip connections that add the input to the sublayer output. This promotes better gradient flow during backpropagation and allows the network to learn perturbations around the identity mapping \cite{vaswani2017attention}.
\end{itemize}

The final output is a sequence of the same shape as the input but enhanced with contextual information learned via attention and local nonlinear transformations from the FFN. Algorithm 1 shows the data flow in the encoder module.

\begin{algorithm}
\caption{Transformer Encoder}
\begin{algorithmic}[1]
\Require Input tensor $X \in \mathbb{R}^{B \times \tau+G \times N}$, where $B$: batch size, $\tau$+G: temporal length and global token size, $N$: number of features
\Ensure Output tensor $Y \in \mathbb{R}^{B \times \tau+G \times N}$

\State Normalize input: $X_{\text{norm}} \leftarrow \text{LayerNorm}(X)$
\State Compute self-attention: $A, \_ \leftarrow \text{MultiHeadAttention}(Q=X_{\text{norm}}, K=X_{\text{norm}}, V=X_{\text{norm}})$
\State Apply dropout and residual connection: $X' \leftarrow X + \text{Dropout}(A)$

\State Normalize: $X'_{\text{norm}} \leftarrow \text{LayerNorm}(X')$
\State Feed-forward network: $F \leftarrow \text{FFN}(X'_{\text{norm}})$
\State Add residual: $Y \leftarrow X' + F$

\State \Return $Y$
\end{algorithmic}
\end{algorithm}

\subsection{GCTAF Framework}

GCTAF utilizes a hierarchical deep learning framework designed to perform solar flare prediction from MVTS input $mvts^{(k)} \in \mathbb{R}^{\tau \times N}$. The framework integrates both local temporal dynamics and global contextual interactions through a dual-stage attention strategy. Specifically, a set of learnable global tokens is introduced to capture long-range dependencies via cross-attention, while the input sequence itself is processed through a series of transformer-based encoder modules for localized modeling. The pooled representations from local and global branches are subsequently fused and passed through a final multilayer perceptron (MLP) layer to perform the classification. The framework components are described below:

\begin{itemize}[leftmargin=*]
    \item \textbf{Learnable global tokens:} A fixed set of $G$ trainable embeddings is initialized to serve as global context vectors. These tokens are intended to distill high-level, class-relevant representations across all time steps, capturing abstract signatures associated.

    \item \textbf{Cross-attention layer:} The global tokens attend to the whole input sequence using multi-head attention. This enables model to selectively extract temporal patterns from the MVTS input that are most informative for the global tokens, thereby enhancing their semantic richness. Technically, this involves projecting both tokens and inputs into query-key-value spaces and aggregating based on attention weights.

    \item \textbf{Fusion via concatenation:} After cross-attention, the enriched global tokens are concatenated with original input sequence along temporal axis. This augmented sequence comprises both raw temporal observations and globally-aware embeddings, promoting bidirectional context exchange.

    \item \textbf{Transformer modules:} A stack of $L$ transformer-based encoder modules (as described previously) is applied to the fused sequence. These blocks perform local self-attention across the extended temporal dimension, allowing joint reasoning over both fine-grained and fused global information.

    \item \textbf{Token separation and pooling:} Once processed, sequence is split back into local and global subsets. Local tokens (original sequence positions) undergo adaptive average pooling along time axis to create a compact representation of short- and mid-range patterns. Simultaneously, global tokens are mean-aggregated across the token dimension to yield a holistic summary.

    \item \textbf{MLP classification head:} The concatenated pooled vector $[\text{Local}, \text{Global}] \in \mathbb{R}^{2N}$ is passed through an MLP head to transform the fused representation into a space for class predictions.

    \item \textbf{Output layer:} A final linear projection maps the learned representations to the desired number of output classes (e.g., F, NF), completing the classification pipeline.
\end{itemize}

Algorithm 2 shows the data flow in the framework.

\begin{algorithm}
\caption{GCTAF Classifier Model}
\begin{algorithmic}[1]
\Require Input sequence $X \in \mathbb{R}^{B \times \tau \times N}$, number of global tokens $G$, number of Transformer blocks $L$
\Ensure Output class probabilities $\hat{y} \in \mathbb{R}^{B \times C}$, where $C$: number of classes

\State Expand global tokens across batch: $G' \leftarrow \text{expand}(G, B)$
\State Perform Cross-Attention: $G_{\text{attn}} \leftarrow \text{MultiHeadAttention}(Q=G', K=X, V=X)$
\State Concatenate local and global features: $X' \leftarrow \text{concat}(X, G_{\text{attn}}, \text{dim}=1)$

\For{$i = 1$ to $L$}
    \State $X' \leftarrow \text{TransformerEncoderBlock}(X')$
\EndFor

\State Separate local and global tokens:
\State \hspace{1em} $X_{\text{local}} \leftarrow X'[:, :-G, :]$
\State \hspace{1em} $X_{\text{global}} \leftarrow X'[:, -G:, :]$

\State Pool features:
\State \hspace{1em} $v_{\text{local}} \leftarrow \text{AvgPool1D}(X_{\text{local}})$
\State \hspace{1em} $v_{\text{global}} \leftarrow \text{Mean}(X_{\text{global}}, \text{dim}=1)$

\State Concatenate pooled vectors: $v \leftarrow \text{concat}(v_{\text{local}}, v_{\text{global}})$

\State Pass through MLP head: $z \leftarrow \text{MLP}(v)$

\State Compute logits: $\hat{y} \leftarrow \text{Linear}(z)$

\State \Return $\hat{y}$
\end{algorithmic}
\end{algorithm}

\section{Experimental Evaluation} \label{sec:Experimental Evaluation}
This section presents our experimental results. The source code is available on our GitHub repository\footnote{\url{https://github.com/OnurVural/gctaf}}.

\subsection{Performance Evaluation Metrics}  \label{sec:eval}

To ensure domain-relevant and fair evaluation, it is crucial to employ metrics suited to the characteristics of the SWAN-SF dataset, particularly its severe class imbalance. In such cases, overall accuracy—reflecting only the proportion of correct predictions—offers limited insight into true model performance. Therefore, we adopt three complementary metrics widely used in solar flare prediction literature. The Heidke Skill Score (HSS2) measures improvement over random prediction, while the Gilbert Skill Score (GS) accounts for chance-level flare detections. The True Skill Statistic (TSS), identified as the most reliable indicator under imbalance \cite{bobra2015solar}, quantifies the difference between true and false positive rates, ranging from -1 (no skill) to 1 (perfect skill) \cite{hamdi2017time, bobra2015solar, vural2025solar}.

\subsection{Preprocessing}

We apply a targeted selection strategy when forming training sets: only NF-class samples from the FQ category are retained for model training \cite{vural2024excon}. This approach is motivated by the observation that NF examples in B and C categories often exhibit magnetic field patterns similar to M- and X-class flares \cite{ahmadzadeh2021train}, which can lead to misinterpretation of NF instances resembling true flare signatures. After defining the training set, missing values in the MVTS sequences are addressed using fast Pearson correlation-based k-nearest neighbors (FPCKNN) \cite{eskandarinasab2024impacts}, preserving temporal continuity and inter-feature dependencies. Finally, all data points are Z-score normalized to standardize magnetic field parameters across time.

\subsection{Training and Evaluation Protocol}

To ensure temporal coverage, we consider the SWAN-SF dataset in five temporal partitions (i.e., $P_1$–$P_5$) and use consecutive partitions for training and testing the GCTAF framework. This strict chronological separation ensures model evaluation on genuinely unseen future data, thereby simulating realistic forecasting conditions. For instance, partition $P_1$ covers the period from 2010 to 2012, while $P_2$ spans 2012 to 2014. Training and validation (i.e., from each training partition, 20\% of the data was set aside as a validation subset to monitor training progress and select the best-performing model) on data from 2010–2012 and testing on 2012–2014 mirrors the natural flow of time, thus enhancing temporal integrity of evaluation. In contrast, disregarding temporal coherence by mixing data points from disparate periods during training and testing can result in information leakage, leading to inflated performance metrics and inaccurate model generalizability assessment \cite{ahmadzadeh2021train, alshammari2024transformer}. The four train-test splits used in GCTAF’s experiments are as follows: $P_{1}$-$P_{2}$, $P_{2}$-$P_{3}$, $P_{3}$-$P_{4}$, and $P_{4}$-$P_{5}$.

\subsection{Hyperparameter Sensitivity Analysis}

\begin{figure}[h]
\centering
\includegraphics[width = 0.79\linewidth]{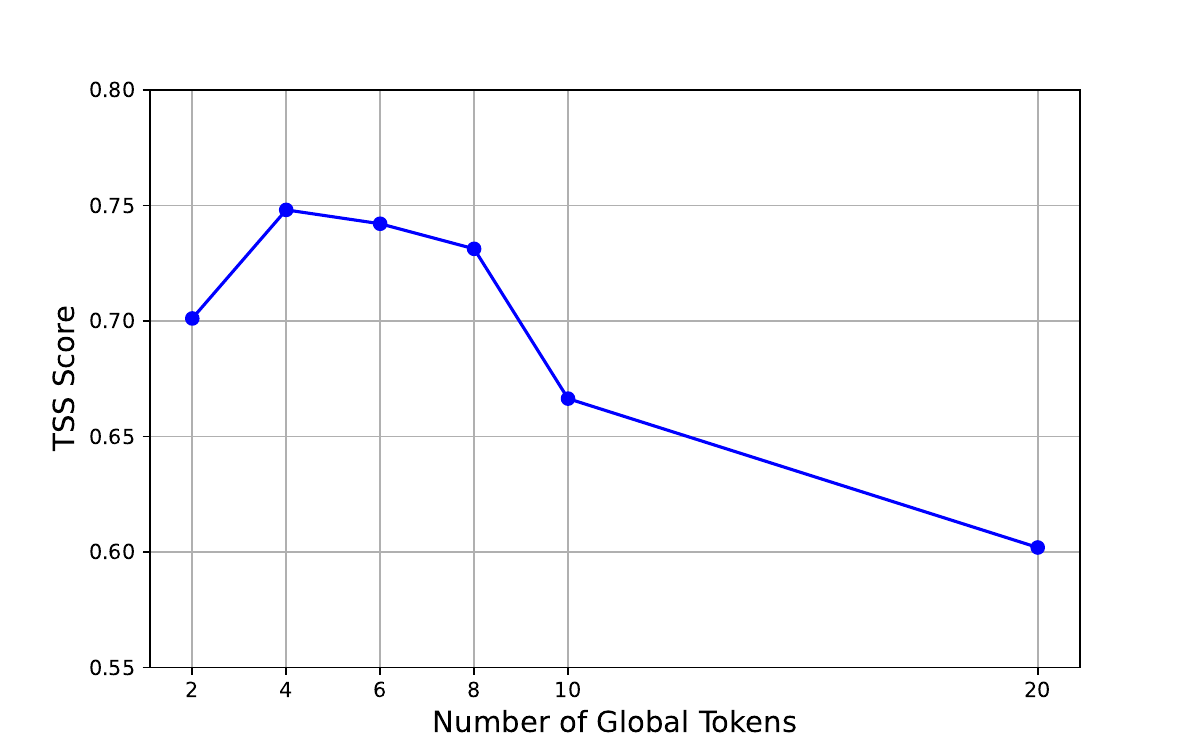}
\caption{Effect of number of global tokens on TSS score.}
\label{fig:numberglobaltoken}
\end{figure}

\begin{figure}[h]
\centering
\includegraphics[width = 0.85\linewidth]{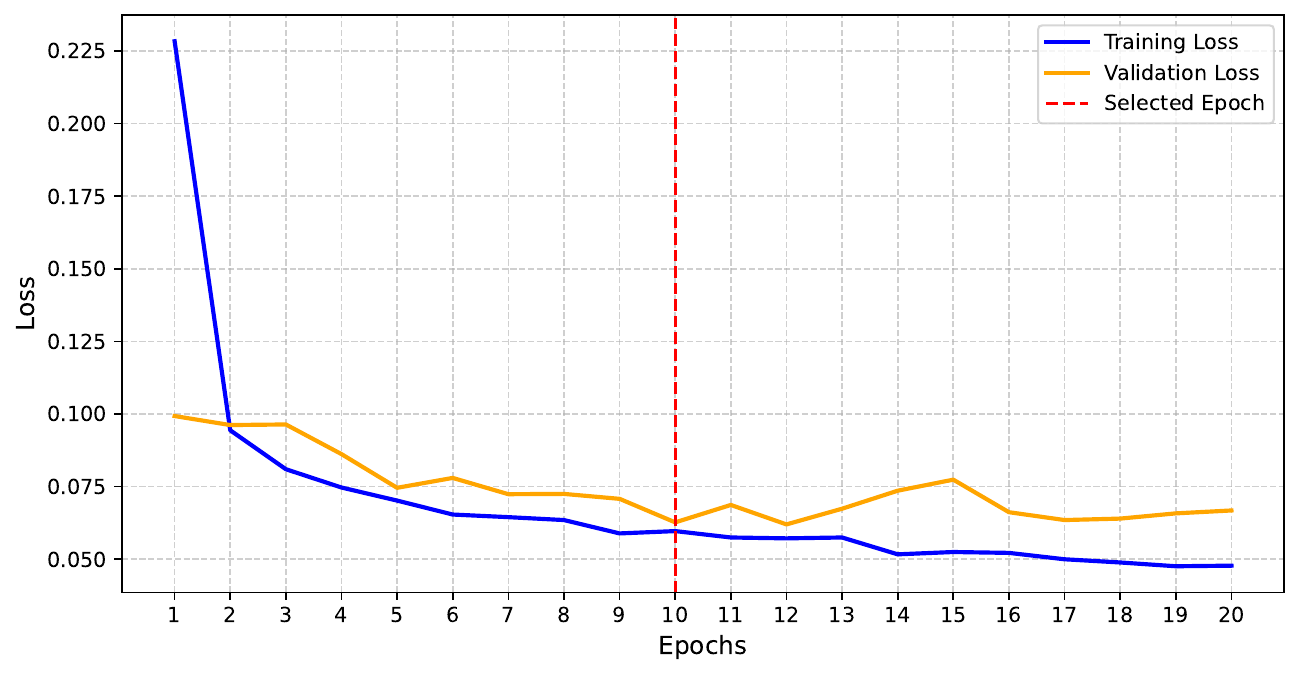}
\caption{Training and validation loss curves for GCTAF over 20 epochs.}
\label{fig:losstrainval}
\end{figure}

We explored how GCTAF’s performance is influenced by hyperparameter choices. We varied the number of transformer encoder modules (1, 5, 10), learning rates (0.0001, 0.001, 0.01), head sizes (64, 128, 256), number of attention heads (2, 4, 8), feedforward dimensions (4, 16, 64), MLP units ([64, 32], [128, 64], [256, 128], [512, 256]), and dropout rates (0.0, 0.1, 0.2) to assess their impact on training convergence, model capacity, and generalization. Table~\ref{tab:hyperparams_tss} gives example configurations and TSS scores from $P_1$ train-validation set. After fixing other parameters, we investigated the effect of global token number on performance. Results in Fig.~\ref{fig:numberglobaltoken} suggest that increasing global tokens initially improves predictions by enabling richer context representation, but beyond a threshold, adding more may reduce performance due to overfitting or increased complexity. Based on these findings, our final configuration is: head size = 256, heads = 4, feedforward dimension = 4, MLP units = [128, 64], global tokens = 4, dropout = 0.1, learning rate = $10^{-4}$. Training epochs were limited, as Fig.~\ref{fig:losstrainval} shows, to prevent overfitting, since extended training had negligible loss improvement but degraded validation performance due to imbalanced temporal partitions.

\begin{table}[t]
\centering
\caption{Hyperparameter configurations and associated TSS scores.}
\label{tab:hyperparams_tss}
\small
\begin{tabular}{|c|c|c|c|c|c|c|}
\hline
\textbf{Head} & \textbf{\#Heads} & \textbf{FF} & \textbf{MLP} & \textbf{\#Global} & \textbf{Dropout} & \textbf{TSS} \\
\textbf{Size} & & \textbf{Dim} & \textbf{Units} & \textbf{Tokens} & & \textbf{Score} \\
\hline
128 & 2 & 4 & [256, 128] & 2 & 0.2 & 0.6911 \\
64 & 8 & 16 & [512, 256] & 6 & 0.1 & 0.7202 \\
256 & 2 & 4 & [128, 64] & 8 & 0.1 & 0.7013 \\
256 & 4 & 4 & [128, 64] & 10 & 0.1 & 0.6664 \\
256 & 4 & 4 & [128, 64] & 4 & 0.1 & 0.7481 \\
\hline
\end{tabular}
\end{table}

\subsection{Ablation Study of Components in GCTAF}

\begin{figure}[h]
\centering
\includegraphics[width = 0.79\linewidth]{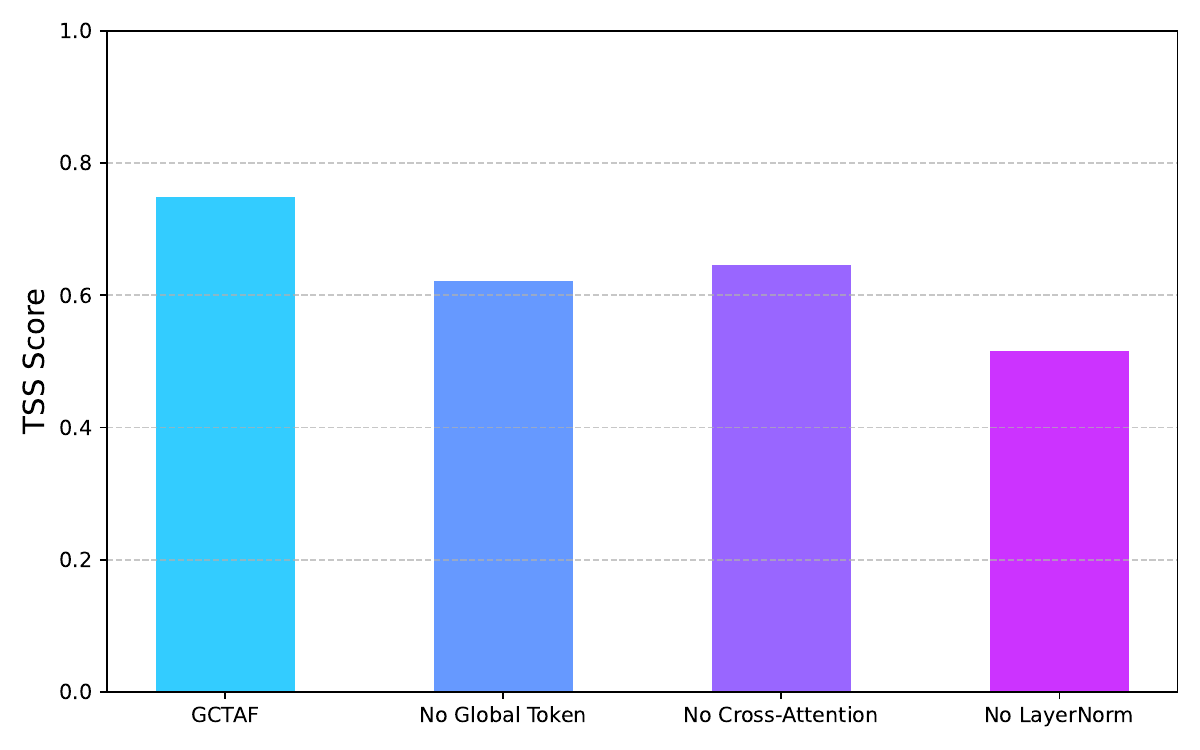}
\caption{Mean TSS results for variants of GCTAF.}
\label{fig:ablation}
\end{figure}

To investigate the contribution of each core component in GCTAF architecture, we conducted an ablation study. Specifically, we tested variants of our model by systematically disabling following modules: (1) no global tokens, where learnable global tokens were excluded from the architecture; (2) no cross-attention, where global tokens were present but not allowed to attend to the input sequence; and (3) no layer normalization, where LayerNorm was removed from transformer modules. The impact of each ablation was assessed based on mean TSS score obtained from four train-test pairs. The results in Fig.~\ref{fig:ablation} indicate the role of each architectural component in achieving optimal flare prediction performance.

\subsection{Baselines}
We selected six baseline methods from contemporary solar flare research. Unless stated otherwise, we used logistic regression for downstream classification to maintain consistency.

\begin{itemize}[leftmargin=*]

\item \textbf{Transformer (TR):} A generic transformer encoder models MVTS data using self-attention mechanisms to capture extended temporal dependencies among magnetic field parameters \cite{alshammari2024transformer}. We used the same hyperparameters in GCTAF.

\item \textbf{Random convolutional kernel transform (ROCKET):} ROCKET classifies time series by convolving each $mvts^{\text{(k)}} \in \mathbb{R}^{\tau \times N}$ with random kernels to extract temporal features for classification \cite{dempster2020rocket}. We used 500 kernels.

\item \textbf{Long short-term memory (LSTM):} LSTM cells process MVTS instances as sequences of timestamp vectors $x^{<t>} \in \mathbb{R}^N$, with the final hidden state encoding the sequence representation for a dense-layer classification \cite{muzaheed2021sequence}. Training minimizes cross-entropy loss for 30 epochs with 128 hidden units, \texttt{tanh} activation, and a learning rate of $10^{-3}$.

\item \textbf{Extreme instance-based contrastive representation learning (EXCON):} EXCON comprises (1) canonical MVTS feature extraction, (2) selection of contrastive representations, (3) temporal embedding via extreme reconstruction loss, and (4) downstream embedding classification \cite{vural2024excon}. Hyperparameters follow the original configuration.

\item \textbf{Support vector machine (SVM):} SVMs are used for flare prediction by learning boundaries from magnetic field features. Each $mvts^{\text{(k)}} \in \mathbb{R}^{\tau \times N}$ is flattened into a vector to train the SVM classifier \cite{bobra2015solar}.

\item \textbf{Vector of last timestamp (VLT):} Magnetic field parameters are extracted at the final time step of the MVTS instance, $x^{<t>} \in \mathbb{R}^N$, representing the state closest to the flare. These vectors are used for downstream classification \cite{bobra2015solar, hamdi2017time}.

\end{itemize}

\subsection{Performance in Solar Flare Prediction}

\begin{figure}[h]
\centering
\includegraphics[width = 0.95\linewidth]{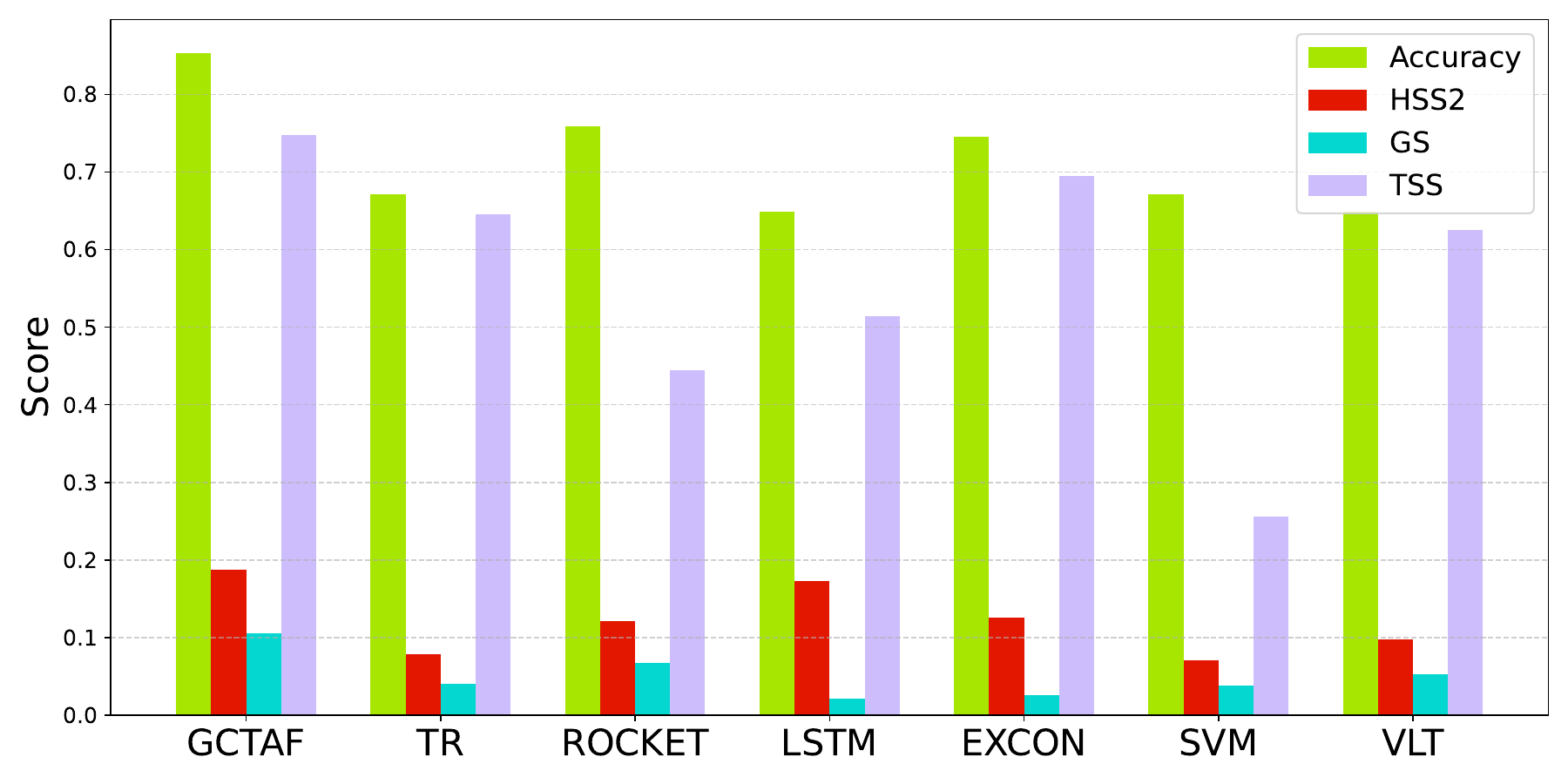}
\caption{Bar chart showing flare prediction scores across four metrics.}
\label{fig:barchart_baselines}
\end{figure}

Here, we evaluate the binary classification performance of GCTAF for predicting solar flares. After running experiments for four train-test partition pairs, the mean performance results in Table~\ref{tab:model_comparison} and Fig.~\ref{fig:barchart_baselines} indicate that GCTAF shows the leading performance in all four metrics against baseline approaches. GCTAF achieves 9.5\% accuracy improvement over ROCKET, the second-best performer. In tasks of such nature, as discussed in Section~\ref{sec:eval}, there is a chance of this high accuracy to be misleading due to potential NF-class overfitting. However, with other metrics, we can verify that is not the case for GCTAF. GCTAF enhances the performance by improving HSS2 1.46\% against LSTM, GS 3.825\% against ROCKET, and TSS 5.3\% against EXCON respectively. Collectively, the results demonstrate that refining transformer-based architectures holds potential in MVTS-driven solar flare prediction tasks as GCTAF is able to produce highly competitive results against state-of-the-art approaches in the literature.  

\setlength{\tabcolsep}{2pt}
\begin{table}[t]
\caption{Solar flare prediction  performance summary (mean ± std)}
\label{tab:model_comparison}
\hspace*{-0.4cm}
\centering
\footnotesize
\begin{tabular}{@{}lcccc@{}}
\toprule
\textbf{Model} & \textbf{Accuracy} & \textbf{HSS2} & \textbf{GS} & \textbf{TSS} \\
\midrule
GCTAF & \textbf{0.8537 ± 0.0472} & \textbf{0.1869 ± 0.0960} & \textbf{0.1054 ± 0.0591} & \textbf{0.7481 ± 0.0254} \\
TR    & 0.6713 ± 0.1373 & 0.07795 ± 0.0282 & 0.04072 ± 0.0153 & 0.6450 ± 0.1416 \\
ROCKET & 0.7587 ± 0.2552 & 0.1215 ± 0.1047 & 0.06715 ± 0.0598 & 0.4445 ± 0.2716 \\
LSTM  & 0.6489 ± 0.2783 & 0.1723 ± 0.1690 & 0.02174 ± 0.0079 & 0.5142 ± 0.1891 \\
EXCON & 0.7452 ± 0.0923 & 0.1254 ± 0.0511 & 0.02567 ± 0.0069 & 0.6951 ± 0.1013 \\
SVM   & 0.6708 ± 0.4062 & 0.07103 ± 0.0602 & 0.03759 ± 0.0327 & 0.2563 ± 0.2280 \\
VLT   & 0.6484 ± 0.2484 & 0.09758 ± 0.0736 & 0.05235 ± 0.0413 & 0.6250 ± 0.2494 \\
\bottomrule
\end{tabular}
\end{table}

\section{Conclusion} \label{sec:Conclusion}
In this work, we proposed and evaluated a novel transformer-based method for MVTS-driven solar flare prediction. The aim was to leverage both high-level global patterns and fine-grained local dynamics of MVTS instances for richer sequence representations. GCTAF operated by combining learnable global tokens, which attend to the full sequence via cross-attention, with local features from transformer encoders. The high performance on the benchmark dataset highlights the potential of advanced transformer-based architectures as a promising direction for future research in solar flare prediction. Future work could concentrate on enhancing the stability of the transformer architecture through architecture-aware regularization, complemented by solar flare domain-specific data augmentation strategies. Another important direction will be to evaluate the generalizability of GCTAF across diverse time series datasets, thereby establishing a more definitive understanding of its robustness and broader applicability.

\section*{Acknowledgment}

This project has been supported in part by NSF awards \#2204363, \#2240022, \#2301397, and \#2530946. 

\bibliographystyle{IEEEtran}
\bibliography{bib_onur_bigdata_25.bib}

\end{document}